\documentclass{article}

\usepackage{times}
\usepackage{graphicx}
\usepackage{subfigure}
\usepackage{natbib}
\usepackage{algorithm}
\usepackage{algorithmic}
\usepackage{hyperref}

\usepackage[accepted]{icml2017}
\usepackage{amsmath}
\usepackage{amssymb}
\usepackage{booktabs}
\usepackage{float}
\usepackage{multibib}
\usepackage{enumitem}
\RequirePackage{fixltx2e}
\usepackage[utf8]{inputenc}
\definecolor{mygray}{gray}{0.5}
\definecolor{mygray2}{gray}{0.6}

\renewcommand{\algorithmiccomment}[1]{\;\;\;\textcolor{mygray}{$//$\;\textit{#1}}}

\newcommand{\sq}[2][0]{
        \mbox{$\medmuskip=#1mu\displaystyle#2$}%
  }

\newcites{main,appendix}{References,References}

\usepackage[LGR,T1]{fontenc} 

\DeclareSymbolFont{upgreek}{LGR}{cmr}{m}{n}
\DeclareMathSymbol{\upsigma}{\mathord}{upgreek}{`s}

\DeclareMathOperator{\softmax}{softmax}

\makeatletter
\newcommand{\algorithmicbreak}{\textbf{break}}
\newcommand{\BREAK}{\STATE \algorithmicbreak}
\makeatother

\usepackage{multirow} 

\usepackage{cleveref}
\icmltitlerunning{Online and Linear-Time Attention by Enforcing Monotonic Alignments}

\usepackage{widetext}

\makeatletter
\renewcommand{\paragraph}{%
  \@startsection{paragraph}{4}%
  {\z@}{0.25ex \@plus 0ex \@minus .2ex}{-1em}%
  {\normalfont\normalsize\bfseries}%
}
\makeatother

\begin{document}

\twocolumn[
        \icmltitle{Online and Linear-Time Attention by Enforcing Monotonic Alignments}

        \icmlsetsymbol{equal}{*}

        \begin{icmlauthorlist}
        \icmlauthor{Colin Raffel}{brain}
        \icmlauthor{Minh-Thang Luong}{brain}
        \icmlauthor{Peter J.\ Liu}{brain}
        \icmlauthor{Ron J.\ Weiss}{brain}
        \icmlauthor{Douglas Eck}{brain}
        \end{icmlauthorlist}

        \icmlaffiliation{brain}{Google Brain, Mountain View, California, USA}

        \icmlcorrespondingauthor{Colin Raffel}{craffel@gmail.com}

        \icmlkeywords{recurrent neural networks, attention mechanism, sequence to sequence}

        \vskip 0.3in
]

\printAffiliationsAndNotice{}

\begin{abstract}
Recurrent neural network models with an attention mechanism have proven to be extremely effective on a wide variety of sequence-to-sequence problems.
However, the fact that soft attention mechanisms perform a pass over the entire input sequence when producing each element in the output sequence precludes their use in online settings and results in a quadratic time complexity.
Based on the insight that the alignment between input and output sequence elements is monotonic in many problems of interest, we propose an end-to-end differentiable method for learning monotonic alignments which, at test time, enables computing attention online and in linear time.
We validate our approach on sentence summarization, machine translation, and online speech recognition problems and achieve results competitive with existing sequence-to-sequence models.
\end{abstract}

\section{Introduction}
\label{sec:introduction}

Recently, the ``sequence-to-sequence'' framework \citemain{sutskever2014sequence,cho2014learning} has facilitated the use of recurrent neural networks (RNNs) on sequence transduction problems such as machine translation and speech recognition.
In this framework, an input sequence is processed with an RNN to produce an ``encoding''; this encoding is then used by a second RNN to produce the target sequence.
As originally proposed, the encoding is a single fixed-length vector representation of the input sequence.
This requires the model to effectively compress all important information about the input sequence into a single vector.
In practice, this often results in the model having difficulty generalizing to longer sequences than those seen during training \citemain{bahdanau2014neural}.

An effective solution to these shortcomings are attention mechanisms \citemain{bahdanau2014neural}.
In a sequence-to-sequence model with attention, the encoder produces a sequence of hidden states (instead of a single fixed-length vector) which correspond to entries in the input sequence.
The decoder is then allowed to refer back to any of the encoder states as it produces its output.
Similar mechanisms have been used as soft addressing schemes in memory-augmented neural network architectures \citemain{graves2014neural,sukhbaatar2015end} and RNNs used for sequence generation \citemain{graves2013generating}.
Attention-based sequence-to-sequence models have proven to be extremely effective on a wide variety of problems, including machine translation \citemain{bahdanau2014neural,luong2015effective}, image captioning \citemain{xu2015show}, speech recognition \citemain{chorowski2015attention,chan2016listen}, and sentence summarization \citemain{rush2015neural}.
In addition, attention creates an implicit soft alignment between entries in the output sequence and entries in the input sequence, which can give useful insight into the model's behavior.

A common criticism of soft attention is that the model must perform a pass over the entire input sequence when producing each element of the output sequence.
This results in the decoding process having complexity $\mathcal{O}(TU)$, where $T$ and $U$ are the input and output sequence lengths respectively.
Furthermore, because the entire sequence must be processed prior to outputting any symbols, soft attention cannot be used in ``online'' settings where output sequence elements are produced when the input has only been partially observed.

The focus of this paper is to propose an alternative attention mechanism which has linear-time complexity and can be used in online settings.
To achieve this, we first note that in many problems, the input-output alignment is roughly monotonic.
For example, when transcribing an audio recording of someone saying ``good morning'', the region of the speech utterance corresponding to ``good'' will always precede the region corresponding to ``morning''.
Even when the alignment is not strictly monotonic, it often only contains local input-output reorderings.
Separately, despite the fact that soft attention allows for assignment of focus to multiple disparate entries of the input sequence, in many cases the attention is assigned mostly to a single entry.
For examples of alignments with these characteristics, we refer to e.g.\ (\citealt{chorowski2015attention} Figure 2; \citealt{chan2016listen} Figure 2; \citealt{rush2015neural} Figure 1; \citealt{bahdanau2014neural} Figure 3), etc.
Of course, this is not true in all problems; for example, when using soft attention for image captioning, the model will often change focus arbitrarily between output steps and will spread attention across large regions of the input image \citemain{xu2015show}.

Motivated by these observations, we propose using \textit{hard monotonic} alignments for sequence-to-sequence problems because, as we argue in \cref{sec:online_process}, they enable computing attention online and in linear time.
Towards this end, we show that it is possible to train such an attention mechanism with a quadratic-time algorithm which computes its expected output.
This allows us to continue using standard backpropagation for training while still facilitating efficient online decoding at test-time.
On all problems we studied, we found these added benefits only incur a small decrease in performance compared to $\softmax$-based attention.

The rest of this paper is structured as follows:
In the following section, we develop an interpretation of soft attention as optimizing a stochastic process in expectation and formulate a corresponding stochastic process which allows for online and linear-time decoding by relying on hard monotonic alignments.
In analogy with soft attention, we then show how to compute the expected output of the monotonic attention process and elucidate how the resulting algorithm differs from standard $\softmax$ attention.
After giving an overview of related work, we apply our approach to the tasks of sentence summarization, machine translation, and online speech recognition, achieving results competitive with existing sequence-to-sequence models.
Finally, we present additional derivations, experimental details, and ideas for future research in the appendix.

\section{Online and Linear-Time Attention}

To motivate our approach, we first point out that $\softmax$-based attention is computing the expected output of a simple stochastic process.
We then detail an alternative process which enables online and linear-time decoding.
Because this process is nondifferentiable, we derive an algorithm for computing its expected output, allowing us to train a model with standard backpropagation while applying our online and linear-time process at test time.
Finally, we propose an alternative energy function motivated by the differences between monotonic attention and $\softmax$-based attention.

\subsection{Soft Attention}

To begin with, we review the commonly-used form of soft attention proposed originally in \citemain{bahdanau2014neural}.
Broadly, a sequence-to-sequence model produces a sequence of outputs based on a processed input sequence.
The model consists of two RNNs, referred to as the ``encoder'' and ``decoder''.
The encoder RNN processes the input sequence $\mathbf{x} = \{x_1, \ldots, x_T\}$  to produce a sequence of hidden states $\mathbf{h} = \{h_1, \ldots, h_T\}$.
We refer to $\mathbf{h}$ as the ``memory'' to emphasize its connection to memory-augmented neural networks \citemain{graves2014neural,sukhbaatar2015end}.
The decoder RNN then produces an output sequence $\mathbf{y} = \{y_1, \ldots, y_U\}$, conditioned on the memory, until a special end-of-sequence token is produced.

When computing $y_i$, a soft attention-based decoder uses a learnable nonlinear function $a(\cdot)$ to produce a scalar value $e_{i, j}$ for each entry $h_j$ in the memory based on $h_j$ and the decoder's state at the previous timestep $s_{i - 1}$.
Typically, $a(\cdot)$ is a single-layer neural network using a $\tanh$ nonlinearity, but other functions such as a simple dot product between $s_{i - 1}$ and $h_j$ have been used \citemain{luong2015effective,graves2014neural}.
These scalar values are normalized using the $\softmax$ function to produce a probability distribution over the memory, which is used to compute a context vector $c_i$ as the weighted sum of $\mathbf{h}$.
Because items in the memory have a sequential correspondence with items in the input, these attention distributions create a soft alignment between the output and input.
Finally, the decoder updates its state to $s_i$ based on $s_{i - 1}$ and $c_i$ and produces $y_i$.
In total, producing $y_i$ involves
\begin{align}
        e_{i, j} &= a(s_{i - 1}, h_j) \label{eq:softmax_energy} \\
        \alpha_{i, j} &= \exp(e_{i, j})\bigg/\sum_{k = 1}^T \exp(e_{i, k}) \label{eq:softmax} \\
        c_i &= \sum_{j = 1}^T \alpha_{i, j} h_j \label{eq:softmax_context} \\
        s_i &= f(s_{i - 1}, y_{i - 1}, c_i) \\
        y_i &= g(s_i, c_i)
\end{align}
where $f(\cdot)$ is a recurrent neural network (typically one or more LSTM \citemain{hochreiter1997long} or GRU \citemain{chung2014empirical} layers) and $g(\cdot)$ is a learnable nonlinear function which maps the decoder state to the output space (e.g.\ an affine transformation followed by a $\softmax$ when the target sequences consist of discrete symbols).

\begin{figure}[t]
\vskip 0.2in
\begin{center}
\centerline{\includegraphics[width=.7\columnwidth]{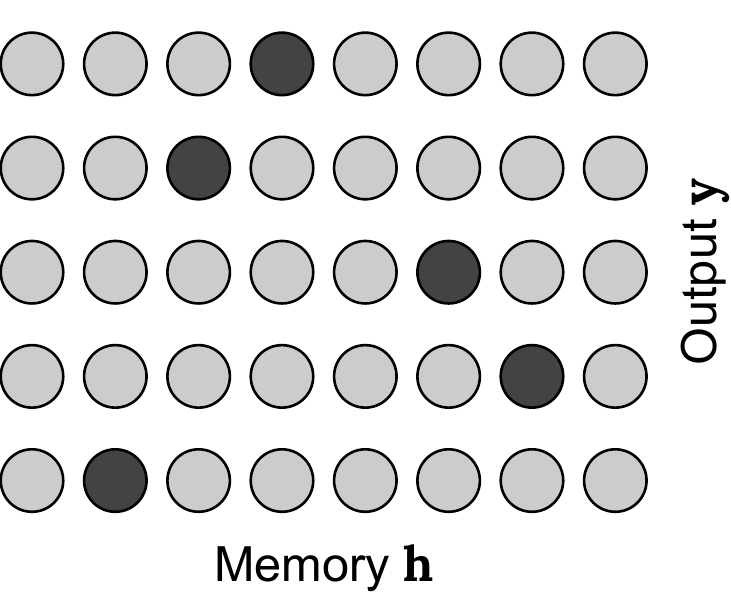}}
\caption{Schematic of the stochastic process underlying $\softmax$-based attention decoders.
Each node represents a possible alignment between an entry of the output sequence (vertical axis) and the memory (horizontal axis).
At each output timestep, the decoder inspects all memory entries (indicated in gray) and attends to a single one (indicated in black).
A black node indicates that memory element $h_j$ is aligned to output $y_i$.
In terms of which memory entry is chosen, there is no dependence across output timesteps or between memory entries.}
\label{fig:attention_process}
\end{center}
\vskip -0.2in
\end{figure}

To motivate our monotonic alignment scheme, we observe that \cref{eq:softmax,eq:softmax_context} are computing the expected output of a simple stochastic process, which can be formulated as follows:
First, a probability $\alpha_{i, j}$ is computed independently for each entry $h_j$ of the memory.
Then, a memory index $k$ is sampled by $k \sim \mathrm{Categorical}(\alpha_i)$ and $c_i$ is set to $h_k$.
We visualize this process in \cref{fig:attention_process}.
Clearly, \cref{eq:softmax_context} shows that soft attention replaces sampling $k$ and assigning $c_i = h_k$ with direct computation of the expected value of $c_i$.

\subsection{A Hard Monotonic Attention Process}
\label{sec:online_process}

The discussion above makes clear that $\softmax$-based attention requires a pass over the entire memory to compute the terms $\alpha_{i, j}$ required to produce each element of the output sequence.
This precludes its use in online settings, and results in a complexity of $\mathcal{O}(TU)$ for generating the output sequence.
In addition, despite the fact that $\textbf{h}$ represents a transformation of a sequence (which ostensibly exhibits dependencies between subsequent elements), the attention probabilities are computed independent of temporal order and the attention distribution at the previous timestep.

\begin{figure}[t]
\vskip 0.2in
\begin{center}
\centerline{\includegraphics[width=.7\columnwidth]{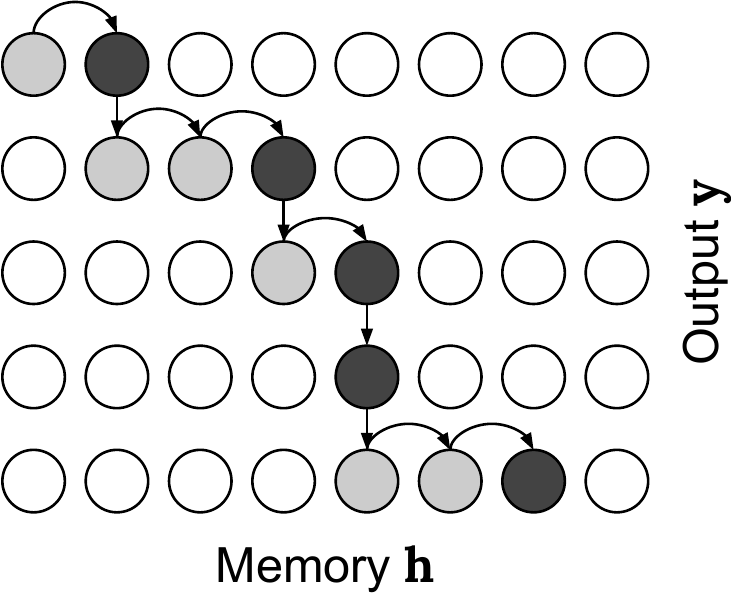}}
\caption{Schematic of our novel monotonic stochastic decoding process.
At each output timestep, the decoder inspects memory entries (indicated in gray) from left-to-right starting from where it left off at the previous output timestep and chooses a single one (indicated in black).
A black node indicates that memory element $h_j$ is aligned to output $y_i$.
White nodes indicate that a particular input-output alignment was not considered because it violates monotonicity.
Arrows indicate the order of processing and dependence between memory entries and output timesteps.}
\label{fig:monotonic_process}
\end{center}
\vskip -0.2in
\end{figure}

We address these shortcomings by first formulating a stochastic process which explicitly processes the memory in a left-to-right manner.
Specifically, for output timestep $i$ we begin processing memory entries from index $t_{i - 1}$, where $t_i$ is the index of the memory entry chosen at output timestep $i$ (for convenience, letting $t_0 = 1$).
We sequentially compute, for $j = t_{i - 1}, t_{i - 1} + 1, t_{i - 1} + 2, \ldots$
\begin{align}
e_{i, j} &= a(s_{i - 1}, h_j) \label{eq:hard_energy} \\
p_{i, j} &= \upsigma(e_{i, j}) \label{eq:hard_sigmoid} \\
z_{i, j} &\sim \mathrm{Bernoulli}(p_{i, j}) \label{eq:z_ij}
\end{align}
where $a(\cdot)$ is a learnable deterministic ``energy function'' and $\upsigma(\cdot)$ is the logistic sigmoid function.
As soon as we sample $z_{i, j} = 1$ for some $j$, we stop and set $c_i = h_j$ and $t_i = j$, ``choosing'' memory entry $j$ for the context vector.
Each $z_{i, j}$ can be seen as representing a discrete choice of whether to ingest a new item from the memory ($z_{i, j} = 0$) or produce an output ($z_{i, j} = 1$).
For all subsequent output timesteps, we repeat this process, always starting from $t_{i - 1}$ (the memory index chosen at the previous timestep).
If for any output timestep $i$ we have $z_{i, j} = 0$ for $j \in \{t_{i - 1}, \ldots, T\}$, we simply set $c_i$ to a vector of zeros.
This process is visualized in \cref{fig:monotonic_process} and is presented more explicitly in \cref{alg:hard_monotonic} (\cref{sec:algorithms}).

Note that by construction, in order to compute $p_{i, j}$, we only need to have computed $h_k$ for $k \in \{1, \ldots, j\}$.
It follows that our novel process can be computed in an online manner; i.e.\ we do not need to wait to observe the entire input sequence before we start producing the output sequence.
Furthermore, because we start inspecting memory elements from where we left off at the previous output timestep (i.e.\ at index $t_{i - 1}$), the resulting process only computes at most $\max(T, U)$ terms $p_{i, j}$, giving it a linear runtime.
Of course, it also makes the strong assumption that the alignment between the input and output sequence is strictly monotonic.

\subsection{Training in Expectation}
\label{sec:softly}

The online alignment process described above involves sampling, which precludes the use of standard backpropagation.
In analogy with $\softmax$-based attention, we therefore propose training with respect to the expected value of $c_i$, which can be computed straightforwardly as follows.
We first compute $e_{i, j}$ and $p_{i, j}$ exactly as in \cref{eq:hard_energy,eq:hard_sigmoid}, where $p_{i, j}$ are interpreted as the probability of choosing memory element $j$ at output timestep $i$.
The attention distribution over the memory is then given by (see \cref{sec:mono_derivation} for a derivation)
\begin{align}
        \alpha_{i, j} &= p_{i, j} \sum_{k = 1}^j \left( \alpha_{i - 1, k} \prod_{l = k}^{j - 1} (1 - p_{i, l}) \right)\label{eq:alpha_ij} \\
                      &= p_{i, j}\left((1 - p_{i, j - 1})\frac{\alpha_{i, j - 1}}{p_{i, j - 1}} + \alpha_{i - 1, j}\right) \label{eq:recurrence_relation}
\end{align}
We provide a solution to the recurrence relation of \cref{eq:recurrence_relation} which allows computing $\alpha_{i, j}$ for $j \in \{1, \ldots, T\}$ in parallel with cumulative sum and cumulative product operations in \cref{sec:recurrence}.
Defining $q_{i, j} = \alpha_{i, j}/p_{i, j}$ gives the following procedure for computing $\alpha_{i, j}$:
\begin{align}
        e_{i, j} &= a(s_{i - 1}, h_j) \label{eq:mono_soft_energy} \\
        p_{i, j} &= \mathrm{\upsigma}(e_{i, j}) \label{eq:mono_sigmoid}\\
        q_{i, j} &= (1 - p_{i, j - 1})q_{i, j - 1} + \alpha_{i - 1, j} \label{eq:mono_q} \\
        \alpha_{i, j} &= p_{i, j}q_{i, j} \label{eq:mono_masked_probs}
\end{align}
where we define the special cases of $q_{i, 0} = 0, p_{i, 0} = 0$ to maintain equivalence with \cref{eq:alpha_ij}.
As in $\softmax$-based attention, the $\alpha_{i, j}$ values produce a weighting over the memory, which are then used to compute the context vector at each timestep as in \cref{eq:softmax_context}.
However, note that $\alpha_{i}$ may not be a valid probability distribution because $\sum_j \alpha_{i, j} \le 1$.
Using $\alpha_{i}$ as-is, without normalization, effectively associates any additional probability not allocated to memory entries to an additional all-zero memory location.
Normalizing $\alpha_i$ so that $\sum_{j = 1}^T \alpha_{i, j} = 1$ has two issues: First, we can't perform this normalization at test time and still achieve online decoding because the normalization depends on $\alpha_{i, j}$ for $j \in \{1, \ldots, T\}$, and second, it would result in a mismatch compared to the probability distribution induced by the hard monotonic attention process which sets $c_i$ to a vector of zeros when $z_{i, j} = 0$ for $j \in \{t_{i - 1}, \ldots, T\}$.

Note that computing $c_i$ still has a quadratic complexity because we must compute $\alpha_{i, j}$ for $j \in \{1, \ldots, T\}$ for each output timestep $i$.
However, because we are training directly with respect to the expected value of $c_i$, we will train our decoders using \cref{eq:mono_soft_energy,eq:mono_sigmoid,eq:mono_q,eq:mono_masked_probs} and then use the online, linear-time attention process of \cref{sec:online_process} at test time.
Furthermore, if $p_{i, j} \in \{0, 1\}$ these approaches are equivalent, so in order for the model to exhibit similar behavior at training and test time, we need $p_{i, j} \approx 0$ or $p_{i, j} \approx 1$.
We address this in \cref{sec:discreteness}.

\subsection{Modified Energy Function}
\label{sec:energy}

While various ``energy functions'' $a(\cdot)$ have been proposed, the most common to our knowledge is the one proposed in \citemain{bahdanau2014neural}:
\begin{equation}
\label{eq:standard_energy}
a(s_{i - 1}, h_j) = v^\top \tanh(W s_{i - 1} + V h_j + b)
\end{equation}
where $W$ and $V$ are weight matrices, $b$ is a bias vector,\footnote{$b$ is occasionally omitted, but we found it often improves performance and only incurs a modest increase in parameters, so we include it.} and $v$ is a weight vector.
We make two modifications to \cref{eq:standard_energy} for use with our monotonic decoder:
First, while the $\softmax$ is invariant to offset,\footnote{That is, $\softmax(e) = \softmax(e + r)$ for any $r \in \mathbb{R}$.} the logistic sigmoid is not.
As a result, we make the simple modification of adding a scalar variable $r$ after the $\tanh$ function, allowing the model to learn the appropriate offset for the pre-sigmoid activations.
Note that \cref{eq:mono_q} tends to exponentially decay attention over the memory because $1 - p_{i, j} \in [0, 1]$; we therefore initialized $r$ to a negative value prior to training so that $1 - p_{i, j}$ tends to be close to 1.
Second, the use of the sigmoid nonlinearity in \cref{eq:mono_sigmoid} implies that our mechanism is particularly sensitive to the scale of the energy terms $e_{i, j}$, or correspondingly, the scale of the energy vector $v$.
We found an effective solution to this issue was to apply weight normalization \citemain{salimans2016weight} to $v$, replacing it by $gv/\|v\|$ where $g$ is a scalar parameter.
Initializing $g$ to the inverse square root of the attention hidden dimension worked well for all problems we studied.

The above produces the energy function
\begin{equation}
\label{eq:modified_energy}
a(s_{i - 1}, h_j) = g\frac{v^\top}{\|v\|} \tanh(W s_{i - 1} + V h_j + b) + r
\end{equation}
The addition of the two scalar parameters $g$ and $r$ prevented the issues described above in all our experiments while incurring a negligible increase in the number of parameters.

\subsection{Encouraging Discreteness}
\label{sec:discreteness}

As mentioned above, in order for our mechanism to exhibit similar behavior when training in expectation and when using the hard monotonic attention process at test time, we require that $p_{i, j} \approx 0$ or $p_{i, j} \approx 1$.
A straightforward way to encourage this behavior is to add noise before the sigmoid in \cref{eq:mono_sigmoid}, as was done e.g.\ in \citemain{frey1997continuous,salakhutdinov2009semantic,foerster2016learning}.
We found that simply adding zero-mean, unit-variance Gaussian noise to the pre-sigmoid activations was sufficient in all of our experiments.
This approach is similar to the recently proposed Gumbel-Softmax trick \citemain{jang2016categorical,maddison2016concrete}, except we did not find it necessary to anneal the temperature as suggested in \citemain{jang2016categorical}.

Note that once we have a model which produces $p_{i, j}$ which are effectively discrete, we can eschew the sampling involved in the process of \cref{sec:online_process} and instead simply set $z_{i, j} = \mathbb{I}(p_{i, j} > \tau)$ where $\mathbb{I}$ is the indicator function and $\tau$ is a threshold.
We used this approach in all of our experiments, setting $\tau = 0.5$.
Furthermore, at test time we do not add pre-sigmoid noise, making decoding purely deterministic.
Combining all of the above, we present our differentiable approach to training the monotonic alignment decoder in \cref{alg:soft_monotonic} (\cref{sec:algorithms}).

\section{Related Work}

\citemain{luo2016learning} and \citemain{zaremba2015reinforcement} both study a similar framework in which a decoder RNN can decide whether to ingest another entry from the input sequence or emit an entry of the output sequence.
Instead of training in expectation, they maintain the discrete nature of this decision while training and use reinforcement learning (RL) techniques.
We initially experimented with RL-based training methods but were unable to find an approach which worked reliably on the different tasks we studied.
Empirically, we also show superior performance to \citemain{luo2016learning} on online speech recognition tasks; we did not attempt any of the tasks from \citemain{zaremba2015reinforcement}.
\citemain{aharoni2016sequence} also study hard monotonic alignments, but their approach requires target alignments computed via a separate statistical alignment algorithm in order to be trained.

As an alternative approach to monotonic alignments, Connectionist Temporal Classification (CTC) \citemain{graves2006connectionist} and the RNN Transducer \citemain{graves2012sequence} both assume that the output sequences consist of symbols, and add an additional ``null'' symbol which corresponds to ``produce no output''.
More closely to our model, \citemain{yu2016online} similarly add ``shift'' and ``emit'' operations to an RNN.
Finally, the Segmental RNN \citemain{kong2015segmental} treats a segmentation of the input sequence as a latent random variable.
In all cases, the alignment path is marginalized out via a dynamic program in order to obtain a conditional probability distribution over label sequences and train directly with maximum likelihood.
These models either require conditional independence assumptions between output symbols or don't condition the decoder (language model) RNN on the input sequence.
We instead follow the framework of attention and marginalize out alignment paths when computing the context vectors $c_i$ which are subsequently fed into the decoder RNN, which allows the decoder to condition on its past output as well as the input sequence.
Our approach can therefore be seen as a marriage of these CTC-style techniques and attention.
Separately, instead of performing an approximate search for the most probable output sequence at test time, we use hard alignments which facilitates linear-time decoding.

A related idea is proposed in \citemain{raffel2017training}, where ``subsampling'' probabilities are assigned to each entry in the memory and a stochastic process is formulated which involves keeping or discarding entries from the input sequence according to the subsampling probabilities.
A dynamic program similar to the one derived in \cref{sec:softly} is then used to compute the expected output which allows for training with standard backpropagation.
Our approach differs in that we utilize an RNN decoder to construct the output sequence, and furthermore allows for output sequences which are longer than the input.

Some similar ideas to those in \cref{sec:softly} were proposed in the context of speech recognition in \citemain{chorowski2015attention}:
First, the prior attention distributions are convolved with a bank of one-dimensional filters and then included in the energy function calculation.
Second, instead of computing attention over the entire memory they only compute it over a sliding window.
This reduces the runtime complexity at the expense of the strong assumption that memory locations attended to at subsequent output timesteps fall within a small window of one another.
Finally, they also advocate replacing the $\softmax$ function with a sigmoid, but they then normalize by the sum of these sigmoid activations across the memory window instead of interpreting these probabilities in the left-to-right framework we use.
While these modifications encourage monotonic attention, they do not explicitly enforce it, and so the authors do not investigate online decoding.

In a similar vein, \citemain{luong2015effective} explore only computing attention over a small window of the memory.
In addition to simply monotonically increasing the window location at each output timestep, they also consider learning a policy for producing the center of the memory window based on the current decoder state.

\citemain{kim2017structured} also make the connection between soft attention and selecting items from memory in expectation.
They consider replacing the $\softmax$ in standard soft attention with an elementwise sigmoid nonlinearity, but do not formulate the interpretation of addressing memory from left-to-right and the corresponding probability distributions as we do in \cref{sec:softly}.

\citemain{jaitly2015neural} apply standard $\softmax$ attention in online settings by splitting the input sequence into chunks and producing output tokens using the attentive sequence-to-sequence framework over each chunk.
They then devise a dynamic program for finding the approximate best alignment between the model output and the target sequence.
In contrast, our ingest/emit probabilities $p_{i, j}$ can be seen as adaptively chunking the input sequence (rather than providing a fixed setting of the chunk size) and we instead train by exactly computing the expectation over alignment paths.

\section{Experiments}
\label{sec:experiments}

To validate our proposed approach for learning monotonic alignments, we applied it to a variety of sequence-to-sequence problems: sentence summarization, machine translation, and online speech recognition.
In the following subsections, we give an overview of the models used and the results we obtained; for more details about hyperparamers and training specifics please see \cref{sec:experiment_details}.
Incidentally, all experiments involved predicting discrete symbols (e.g.\ phonemes, characters, or words); as a result, the output of the decoder in each of our models was fed into an affine transformation followed by a $\softmax$ nonlinearity with a dimensionality corresponding to the number of possible symbols.
At test time, we performed a beam search over $\softmax$ predictions on all problems except machine translation.
All networks were trained using standard cross-entropy loss with teacher forcing against target sequences using the Adam optimizer \citemain{kingma2014adam}.
All of our decoders used the monotonic attention mechanism of \cref{sec:softly} during training to address the hidden states of the encoder.
For comparison, we report test-time results using both the hard linear-time decoding method of \cref{sec:online_process} and the ``soft'' monotonic attention distribution.
We also present the results of a synthetic benchmark we used to measure the potential speedup offered by our linear-time decoding process in \cref{sec:faster}.

\paragraph{Online Speech Recognition}

Online speech recognition involves transcribing the words spoken in a speech utterance in real-time, i.e.\ as a person is talking.
This problem is a natural application for monotonic alignments because online decoding is an explicit requirement.
In addition, this precludes the use of bidirectional RNNs, which degrades performance somewhat \citemain{graves2013speech}.
We tested our approach on two datasets: TIMIT \citemain{garofolo1993darpa} and the Wall Street Journal corpus \citemain{paul1992design}.

Speech recognition on the TIMIT dataset involves transcribing the phoneme sequence underlying a given speech utterance.
Speech utterances were represented as sequences of 40-filter (plus energy) mel-filterbank spectra, computed every 10 milliseconds, with delta- and delta-delta-features.
Our encoder RNN consisted of three unidirectional LSTM layers.
Following \citemain{chan2016listen}, after the first and second LSTM layer we placed time reduction layers which skip every other sequence element.
Our decoder RNN was a single unidirectional LSTM.
Our output $\softmax$ had 62 dimensions, corresponding to the 60 phonemes from TIMIT plus special start-of-sequence and end-of-sequence tokens.
At test time, we utilized a beam search over $\softmax$ predictions, with a beam width of 10.
We report the phone error rate (PER) after applying the standard mapping to 39 phonemes \citemain{graves2013speech}.
We used the standard train/validation/test split and report results on the test set.

\begin{table}[t]
\caption{Phone error rate on the TIMIT dataset for different online methods.}
\label{tab:timit_results}
\vskip 0.15in
\begin{center}
\begin{small}
\begin{tabular}{lc}
\toprule
Method & PER \\
\midrule
\citemain{luo2016learning} (stacked LSTM) & 21.5\% \\
\citemain{jaitly2015neural} (end-to-end) & 20.8\% \\
\citemain{luo2016learning} (grid LSTM) & 20.5\% \\
Hard Monotonic Attention (ours) & 20.4\% \\
Soft Monotonic Attention (ours, offline) & 20.1\% \\
\citemain{graves2013speech} (CTC) & 19.6\% \\
\bottomrule
\end{tabular}
\end{small}
\end{center}
\vskip -0.2in
\end{table}

Our model's performance, with a comparison to other online approaches, is shown in \cref{tab:timit_results}.
We achieve better performance than recently proposed sequence-to-sequence models \citemain{luo2016learning,jaitly2015neural}, though the small size of the TIMIT dataset and the resulting variability of results precludes making substantiative claims about one approach being best.
We note that \citemain{jaitly2015neural} were able to improve performance by precomputing alignments using an HMM system and providing them as a supervised signal to their decoder; we did not experiment with this idea.
CTC \citemain{graves2013speech} still outperforms all sequence-to-sequence models.
In addition, there remains a substantial gap between these online results and offline results using bidirectional LSTMs, e.g. \citemain{chorowski2015attention} achieves a 17.6\% phone error rate using a $\softmax$-based attention mechanism and \citemain{graves2013speech} achieved 17.7\% using a pre-trained RNN transducer model.
We are interested in investigating ways to close this gap in future work.

Because of the size of the dataset, performance on TIMIT is often highly dependent on appropriate regularization.
We therefore also evaluated our approach on the Wall Street Journal (WSJ) speech recognition dataset, which is about 10 times larger.
For the WSJ corpus, we present speech utterances to the network as 80-filter mel-filterbank spectra with delta- and delta-delta features, and normalized using per-speaker mean and variance computed offline.
The model architecture is a variation of that from \citemain{zhang2016very}, using an 8 layer encoder including: two convolutional layers which downsample the sequence in time, followed by one unidirectional convolutional LSTM layer, and finally a stack of three unidirectional LSTM layers interleaved with linear projection layers and batch normalization.
The encoder output sequence is consumed by the proposed online attention mechanism which is passed into a decoder consisting of a single unidirectional LSTM layer followed by a $\softmax$ layer.

Our output $\softmax$ predicted one of 49 symbols, consisting of alphanumeric characters, punctuation marks, and start-of sequence, end-of-sequence, ``unknown'', ``noise'', and word delimiter tokens.
We utilized label smoothing during training \citemain{chorowski2017towards}, replacing the targets at time $y_t$ with a convex weighted combination of the surrounding five labels (full details in \cref{sec:wsj_details}).
Performance was measured in terms of word error rate (WER) on the test set after segmenting the model's predictions according to the word delimiter tokens.
We used the standard dataset split of si284 for training, dev93 for validation, and eval92 for testing.
We did not use a language model to improve decoding performance.

\begin{table}[t]
\caption{Word error rate on the WSJ dataset.  All approaches used a unidirectional encoder; results in {\color{mygray2}grey} indicate offline models.}
\label{tab:wsj_results}
\vskip 0.15in
\begin{center}
\begin{small}
\begin{tabular}{lc}
\toprule
Method & WER \\
\midrule
CTC (our model) & 33.4\% \\
\citemain{luo2016learning} (hard attention) & 27.0\% \\
\citemain{wang2016lookahead} (CTC) & 22.7\% \\
Hard Monotonic Attention (our model) & 17.4\% \\
\color{mygray2}Soft Monotonic Attention (our model) & \color{mygray2}16.5\% \\
\color{mygray2}Softmax Attention (our model) & \color{mygray2}16.0\% \\
\bottomrule
\end{tabular}
\end{small}
\end{center}
\vskip -0.2in
\end{table}

Our results on WSJ are shown in \cref{tab:wsj_results}.
Our model, with hard monotonic decoding, achieved a significantly lower WER than the other online methods.
While these figures show a clear advantage to our approach, our model architecture differed significantly from those of \citemain{luo2016learning,wang2016lookahead}.
We therefore additionally measured performance against a baseline model which was identical to our model except that it used $\softmax$-based attention (which makes it quadratic-time and offline) instead of a monotonic alignment decoder.
This resulted in a small decrease of 1.4\% WER, suggesting that our hard monotonic attention approach achieves competitive performance while being substantially more efficient.
To get a qualitative picture of our model's behavior compared to the $\softmax$-attention baseline, we plot each model's input-output alignments for two example speech utterances in \cref{fig:speech_alignments} (\cref{sec:figures}).
Both models learn roughly the same alignment, with some minor differences caused by ours being both hard and strictly monotonic.

\paragraph{Sentence Summarization}

Speech recognition exhibits a strictly monotonic input-output alignment.
We are interested in testing whether our approach is also effective on problems which only exhibit approximately monotonic alignments.
We therefore ran a ``sentence summarization'' experiment using the Gigaword corpus, which involves predicting the headline of a news article from its first sentence.

Overall, we used the model of \citemain{liu2016text}, modifying it only so that it used our monotonic alignment decoder instead of a soft attention decoder.
Because online decoding is not important for sentence summarization, we utilized bidirectional RNNs in the encoder for this task (as is standard).
We expect that the bidirectional RNNs will give the model local context which may help allow for strictly monotonic alignments.
The model both took as input and produced as output one-hot representations of the word IDs, with a vocabulary of the 200,000 most common words in the training set.
Our encoder consisted of a word embedding matrix (which was initialized randomly and trained as part of the model) followed by four bidirectional LSTM layers.
We used a single LSTM layer for the decoder.
For data preparation and evaluation, we followed the approach of \citemain{rush2015neural}, measuring performance using the ROUGE metric.

Our results, along with the scores achieved by other approaches, are presented in \cref{tab:summarization_results}.
While the monotonic alignment model outperformed existing models by a substantial margin, it fell slightly behind the model of \citemain{liu2016text} which we used as a baseline.
The higher performance of our model and the model of \citemain{liu2016text} can be partially explained by the fact that their encoders have roughly twice as many layers as most models proposed in the literature.

\begin{table}[t]
\caption{ROUGE F-measure scores for sentence summarization on the Gigaword test set of \citemain{rush2015neural}.
\citemain{rush2015neural} reports ROUGE recall scores, so we report the F-1 scores computed for that approach from \citemain{chopra2016abstractive}.
As is standard, we report unigram, bigram, and longest common subsequence metrics as R-1, R-2, and R-L respectively.}
\label{tab:summarization_results}
\vskip 0.15in
\begin{center}
\begin{small}
\begin{tabular}{lccc}
\toprule
Method & R-1 & R-2 & R-L \\
\midrule
\citemain{zeng2016efficient} & 27.82 & 12.74 & 26.01 \\
\citemain{rush2015neural} & 29.76 & 11.88 & 26.96 \\
\citemain{yu2016online} & 30.27 & 13.68 & 27.91 \\
\citemain{chopra2016abstractive} & 33.78 & 15.97 & 31.15 \\
\citemain{miao2016language} & 34.17 & 15.94 & 31.92 \\
\citemain{nallapati2016abstractive} & 34.19 & 16.29 & 32.13 \\
\citemain{yu2016neural} & 34.41 & 16.86 & 31.83 \\
\citemain{suzuki2017cutting} & 36.30 & 17.31 & 33.88 \\
Hard Monotonic (ours) & 37.14 & 18.00 & 34.87 \\
Soft Monotonic (ours) & 38.03 & 18.57 & 35.70 \\
\citemain{liu2016text} & 38.22 & 18.70 & 35.74 \\
\bottomrule
\end{tabular}
\end{small}
\end{center}
\vskip -0.2in
\end{table}

\begin{figure}[t]
\vskip 0.2in
\begin{center}
\centerline{\includegraphics[width=\columnwidth]{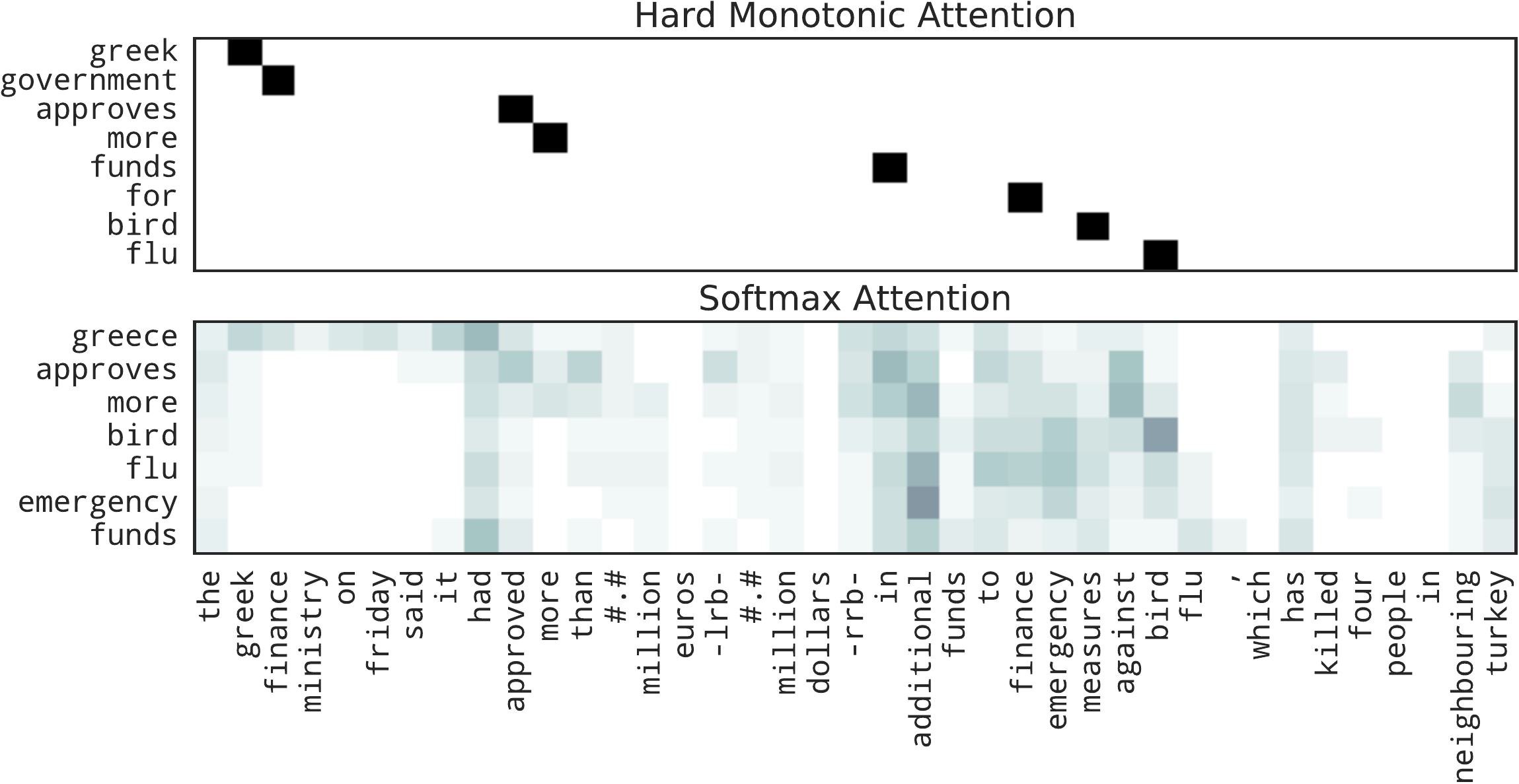}}
\caption{Example sentence-summary pair with attention alignments for our hard monotonic model and the $\softmax$-based attention model of \citemain{liu2016text}.
Attention matrices are displayed so that black corresponds to $1$ and white corresponds to $0$.
The ground-truth summary is ``greece pumps more money and personnel into bird flu defense''.}
\label{fig:summarization_alignments_1}
\end{center}
\vskip -0.4in
\end{figure}

For qualitative evaluation, we plot an example input-output pair and alignment matrices for our hard monotonic attention model and the $\softmax$-attention baseline of \citemain{liu2016text} in \cref{fig:summarization_alignments_1} (an additional example is shown in \cref{fig:summarization_alignments_2}, \cref{sec:figures}).
Most apparent is that a given word in the summary is not always aligned to the most obvious word in the input sentence; the hard monotonic decoder aligns the first four words in the summary reasonably (greek $\leftrightarrow$ greek, government $\leftrightarrow$ finance, approves $\leftrightarrow$ approved, more $\leftrightarrow$ more), but the latter four words have unexpected alignments (funds $\leftrightarrow$ in, to $\leftrightarrow$ for, bird $\leftrightarrow$ measures, bird $\leftrightarrow$ flu).
We believe this is due to the ability of the multilayer bidirectional RNN encoder to reorder words in the input sequence.
This effect is also apparent in \cref{fig:summarization_alignments_2}/ (\cref{sec:figures}), where the monotonic alignment decoder is able to produce the phrase ``human rights criticism'' despite the fact that the input sentence has the phrase ``criticism of human rights''.
Separately, we note that the $\softmax$ attention model's alignments are extremely ``soft'' and non-monotonic; this may be advantageous for this problem and partially explain its slightly superior performance.

\paragraph{Machine Translation}

We also evaluated our approach on machine translation, another task which does not exhibit strictly monotonic alignments.
In fact, for some language pairs (e.g.\ English and Japanese, English and Korean), we do not expect monotonicity at all.
However, for other pairs (e.g.\ English and French, English and Vietnamese) only local word reorderings are required.
Our translation experiments therefore involved English to Vietnamese translation using the parallel corpus of TED talks (133K sentence pairs) provided by the IWSLT 2015 Evaluation Campaign \citemain{iwslt15}.
Following \citemain{luong2015stanford}, we tokenize the corpus with the default Moses tokenizer, preserve casing, and replace words whose frequencies are less than $5$ by \texttt{<unk>}.
As a result, our vocabulary sizes are 17K and 7.7K for English and Vietnamese respectively. We use the TED tst2012 (1553 sentences) as a validation set for hyperparameter tuning and TED tst2013 (1268 sentences) as a test set. We report results in both perplexity and BLEU.

Our baseline neural machine translation (NMT) system is the $\softmax$ attention-based sequence-to-sequence model described in \citemain{luong2015effective}.
From that baseline, we substitute the $\softmax$-based attention mechanism with our proposed monotonic alignment decoder.
The model utilizes two-layer unidirectional LSTM networks for both the encoder and decoder.

In \citemain{luong2015effective}, the authors demonstrated that under their proposed architecture, a dot product-based energy function worked better than \cref{eq:standard_energy}.
Since our architecture is based on that of \citemain{luong2015effective}, to facilitate comparison we also tested the following variant:
\begin{equation}
\label{eq:luong_energy}
a(s_{i - 1}, h_j) = g(s_{i - 1}^{\top}Wh) + r
\end{equation}
where $g$ and $r$ are scalars (initialized as in \cref{sec:energy}) and $W$ is a weight matrix.

Our results are shown in Table~\ref{t:mt_envi}.
To get a better picture of each model's behavior, we plot input-output alignments in \cref{fig:translation_alignments} (\cref{sec:figures}).
Most noticeable is that the monotonic alignment model tends to focus attention later in the input sequence than the baseline $\softmax$-attention model.
We hypothesize that this is a way to compensate for non-monotonic alignments when a unidirectional encoder is used; i.e.\ the model has effectively learned to focus on words at the end of phrases which require reordering, at which point the unidirectional encoder has observed the whole phrase.
This can be seen most clearly in the example on the right, where translating ``a huge famine'' to Vietnamese requires reordering (as suggested by the $\softmax$-attention model's alignment), so the hard monotonic alignment model focuses attention on the final word in the phrase (``famine'') while producing its translation.
We suspect our model's small decrease in BLEU compared to the baseline model may be due in part to this increased modeling burden.

\begin{table}[t]
\caption{Performance on the IWSLT 2015 English-Vietnamese TED talks for our monotonic alignment model and the baseline $\softmax$-attention model of \citemain{luong2015stanford}.}
\label{t:mt_envi}
\vskip 0.15in
\begin{center}
\begin{small}
\begin{tabular}{lc}
\toprule
Method & BLEU \\
\midrule
\citemain{luong2015stanford} & 23.3 \\
Hard Monotonic, energy function \cref{eq:modified_energy} & 22.6 \\
Hard Monotonic, energy function \cref{eq:luong_energy} & 23.0 \\
\bottomrule
\end{tabular}
\end{small}
\end{center}
\vskip -0.2in
\end{table}

\section{Discussion}

Our results show that our differentiable approach to enforcing monotonic alignments can produce models which, following the decoding process of \cref{sec:online_process}, provide efficient online decoding at test time without sacrificing substantial performance on a wide variety of tasks.
We believe our framework presents a promising environment for future work on online and linear-time sequence-to-sequence models.
We are interested in investigating various extensions to this approach, which we outline in \cref{sec:future_work}.
To facilitate experimentation with our proposed attention mechanism, we have made an example TensorFlow \citemain{abadi2016tensorflow} implementation of our approach available online\footnote{\url{https://github.com/craffel/mad}} and added a reference implementation to TensorFlow's \texttt{tf.contrib.seq2seq} module.
We also provide a ``practitioner's guide'' in \cref{sec:practitioners}.

\section*{Acknowledgements}

We thank Jan Chorowski, Mark Daoust, Pietro Kreitlon Carolino, Dieterich Lawson, Navdeep Jaitly, George Tucker, Quoc V.\ Le, Kelvin Xu, Cinjon Resnick, Melody Guan, Matthew D.\ Hoffman, Jeffrey Dean, Kevin Swersky, Ashish Vaswani, and members of the Google Brain team for helpful discussions and insight.

\bibliographystylemain{icml2017}
\bibliographymain{example_paper}

\clearpage

\appendix

\onecolumn

\section{Algorithms}
\label{sec:algorithms}

Below are algorithms for the hard monotonic decoding process we used at test time (\cref{alg:hard_monotonic}) and the approach for computing its expected output that we used to train the network (\cref{alg:soft_monotonic}).
Terminology matches the main text, except we use $\vec{0}$ to signify a vector of zeros.

\begin{algorithm*}[h]
   \caption{Hard Monotonic Attention Process}
   \label{alg:hard_monotonic}
\begin{algorithmic}
   \STATE {\bfseries Input:} memory $\mathbf{h}$ of length $T$
   \STATE {\bfseries State:} $s_0 = \vec{0}, t_0 = 1$, $i = 1$, $y_0 = \mathrm{StartOfSequence}$
   \WHILE[Produce output tokens until end-of-sequence token is produced]{$y_{i - 1} \ne \mathrm{EndOfSequence}$}
   \STATE $\mathrm{finished} = 0$ \COMMENT{Keep track of whether we chose a memory entry or not}
   \FOR[Start inspecting memory entries $h_j$ left-to-right from where we left off]{$j = t_{i - 1}$ {\bfseries to} $T$}
   \STATE $e_{i, j} = a(s_{i - 1}, h_j)$ \COMMENT{\textit{Compute attention energy for $h_j$}}
   \STATE $p_{i, j} = \upsigma(e_{i, j})$ \COMMENT{Compute probability of choosing $h_j$}
   \STATE $z_{i, j} \sim \mathrm{Bernoulli}(p_{i, j})$ \COMMENT{Sample whether to ingest another memory entry or output new symbol}
   \IF[If we sample 1, we stop scanning the memory]{$z_{i, j} = 1$}
   \STATE $c_i = h_j$ \COMMENT{Set the context vector to the chosen memory entry}
   \STATE $t_i = j$ \COMMENT{Remember where we left off for the next output timestep}
   \STATE $\mathrm{finished} = 1$ \COMMENT{Keep track of the fact that we chose a memory entry}
   \BREAK \COMMENT{Stop scanning the memory}
   \ENDIF
   \ENDFOR
   \IF{$\mathrm{finished = 0}$}
   \STATE $c_i = \vec{0}$ \COMMENT{If we scanned the entire memory without selecting anything, set $c_i$ to a vector of zeros}
   \ENDIF
   \STATE $s_i = f(s_{i - 1}, y_{i - 1}, c_i)$ \COMMENT{Update the state based on the new context vector using the RNN $f$}
   \STATE $y_i = g(s_i, c_i)$ \COMMENT{Output a new symbol using the $\softmax$ layer $g$}
   \STATE $i = i + 1$
   \ENDWHILE
\end{algorithmic}
\end{algorithm*}

\setlength{\intextsep}{4pt}

\begin{algorithm*}[h]
   \caption{Soft Monotonic Attention Decoder}
   \label{alg:soft_monotonic}
\begin{algorithmic}
   \STATE {\bfseries Input:} memory $\mathbf{h}$ of length $T$,  target outputs $\mathbf{\hat{y}} = \{\mathrm{StartOfSequence}, \hat{y}_1, \hat{y}_2, \ldots, \mathrm{EndOfSequence}\}$
   \STATE {\bfseries State:} $s_0 = \vec{0}$, $i = 1$, $\alpha_{0, j} = \delta_j$ for $j \in \{1, \ldots, T\}$
   \WHILE[Produce output tokens until end of the target sequence]{$\hat{y}_{i - 1} \ne \mathrm{EndOfSequence}$}
   \STATE $p_{i, 0} = 0, q_{i, 0} = 0$ \COMMENT{Special cases so that the recurrence relation matches \cref{eq:alpha_ij}}
   \FOR[Inspect all memory entries $h_j$]{$j = 1$ {\bfseries to} $T$}
   \STATE $e_{i, j} = a(s_{i - 1}, h_j)$ \COMMENT{Compute attention energy for $h_j$ using \cref{eq:modified_energy}}
   \STATE $e_{i, j} = e_{i, j} + \mathcal{N}(0, 1)$ \COMMENT{Add pre-sigmoid noise to encourage $p_{i, j} \approx 0$ or $p_{i, j} \approx 1$}
   \STATE $p_{i, j} = \upsigma(e_{i, j})$ \COMMENT{Compute probability of choosing $h_j$}
   \STATE $q_{i, j} = (1 - p_{i, j - 1})q_{i, j - 1} + \alpha_{i - 1, j}$ \COMMENT{Iterate recurrence relation derived in \cref{eq:recurrence_relation}}
   \STATE $\alpha_{i, j} = p_{i, j}q_{i, j}$ \COMMENT{Compute the probability that $c_i = h_j$}
   \ENDFOR
   \STATE $c_i = \sum_{j = 1}^T \alpha_{i, j}h_j$ \COMMENT{Compute weighted combination of memory for context vector}
   \STATE $s_i = f(s_{i - 1}, y_{i - 1}, c_i)$ \COMMENT{Update the state based on the new context vector using the RNN $f$}
   \STATE $y_i = g(s_i, c_i)$ \COMMENT{Compute predicted output for timestep $i$ using the $\softmax$ layer $g$}
   \STATE $i = i + 1$
   \ENDWHILE
\end{algorithmic}
\end{algorithm*}

\setlength{\intextsep}{12pt}

\clearpage

\section{Figures}
\label{sec:figures}

Below are example hard monotonic and $\softmax$ attention alignments for each of the different tasks we included in our experiments.  Attention matrices are displayed so that black corresponds to 1 and white corresponds to 0.

\begin{figure*}[h]
\vskip 0.2in
\begin{center}
\centerline{\includegraphics[width=.9\textwidth]{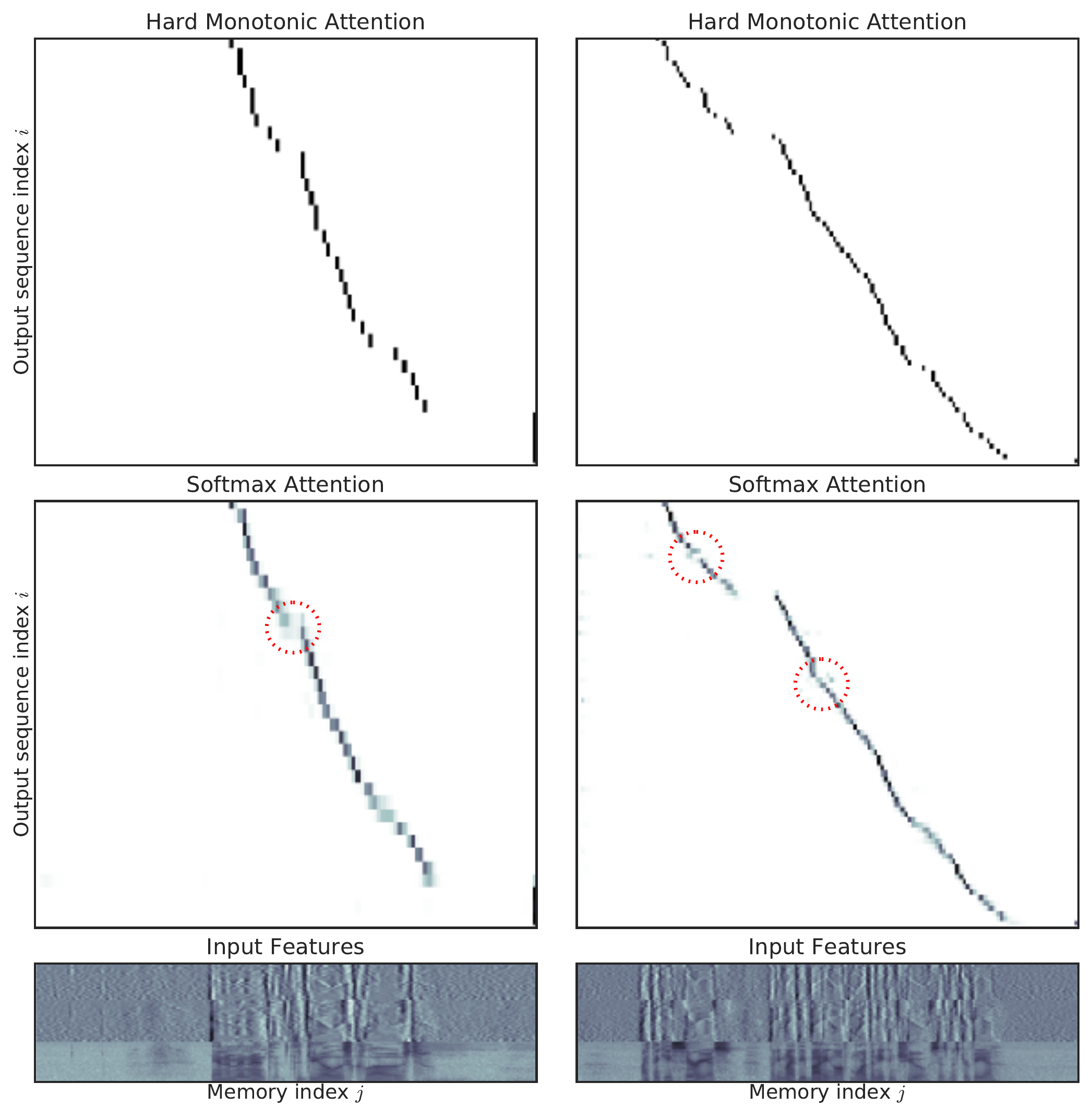}}
\caption{Attention alignments from hard monotonic attention and $\softmax$-based attention models for a two example speech utterances.
From top to bottom, we show the hard monotonic alignment, the $\softmax$-attention alignment, and the utterance feature sequence.
Differences in the alignments are highlighted with dashed red circles.
Gaps in the alignment paths correspond to effectively ignoring silences and pauses in the speech utterances.}
\label{fig:speech_alignments}
\end{center}
\vskip -0.4in
\end{figure*}

\begin{figure*}[p]
\vskip 0.2in
\begin{center}
\centerline{\includegraphics[width=.95\textwidth]{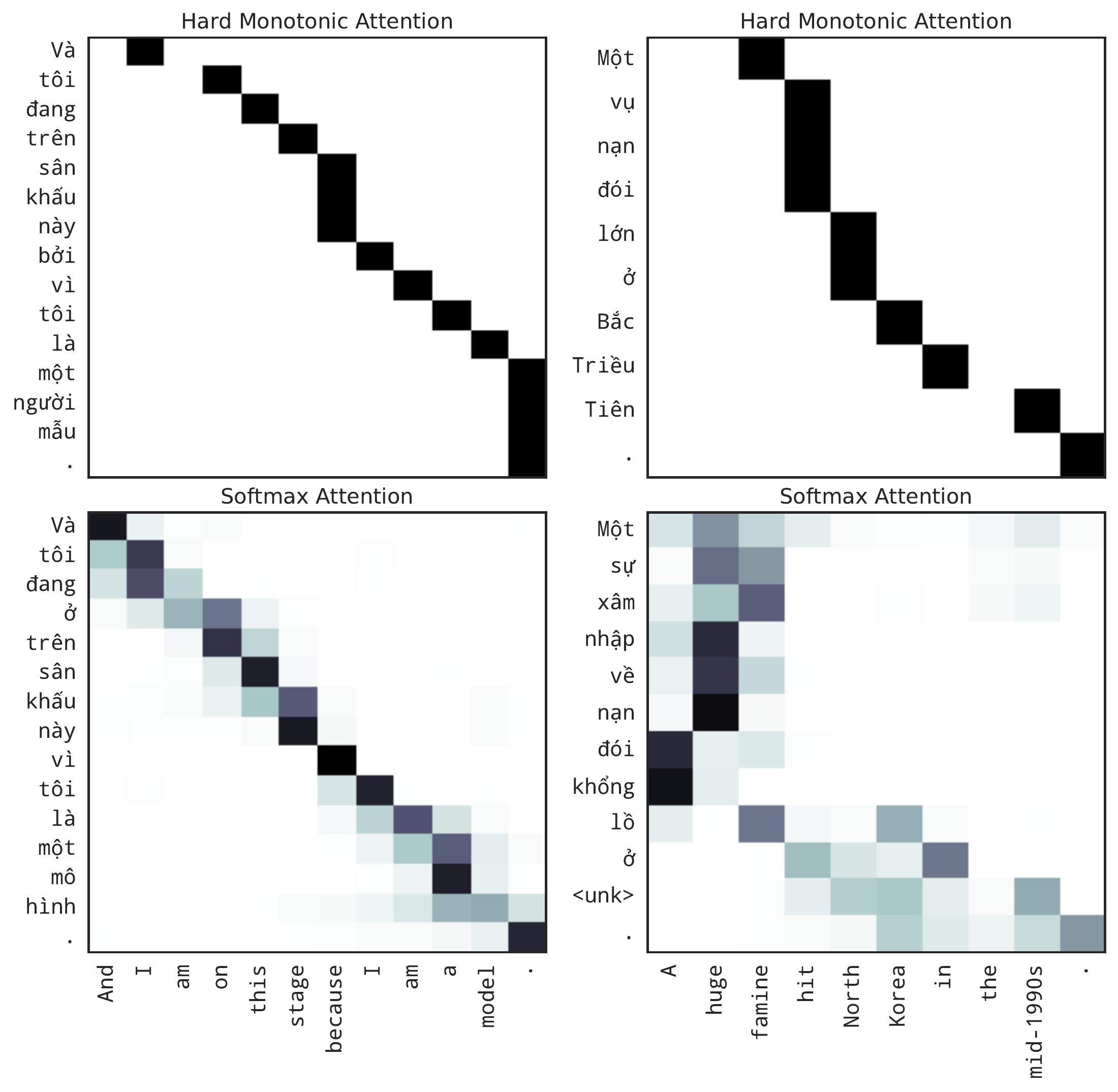}}
\caption{English sentences, predicted Vietnamese sentences, and input-output alignments for our proposed hard monotonic alignment model and the baseline model of \citeappendix{luong2015stanford}.
The Vietnamese model outputs for the left example can be translated back to English as ``And I on this stage because I am a model.'' (monotonic) and ``And I am on this stage because I am a structure.'' ($\softmax$).
The input word ``model'' can mean either a person or a thing; the monotonic alignment model correctly chose the former while the $\softmax$ alignment model chose the latter.
The monotonic alignment model erroneously skipped the first verb in the sentence.
For the right example, translations of the model outputs back to English are ``A large famine in North Korea.'' (monotonic) and ``An invasion of a huge famine in \texttt{<unk>}.'' ($\softmax$).
The monotonic alignment model managed to translate the proper noun North Korea, while the $\softmax$ alignment model produced \texttt{<unk>}.
Both models skipped the phrase ``mid-1990s''; this type of error is common in neural machine translation systems.}
\label{fig:translation_alignments}
\end{center}
\vskip -0.4in
\end{figure*}

\clearpage

\begin{figure*}[h]
\vskip 0.2in
\begin{center}
\centerline{\includegraphics[width=.8\textwidth]{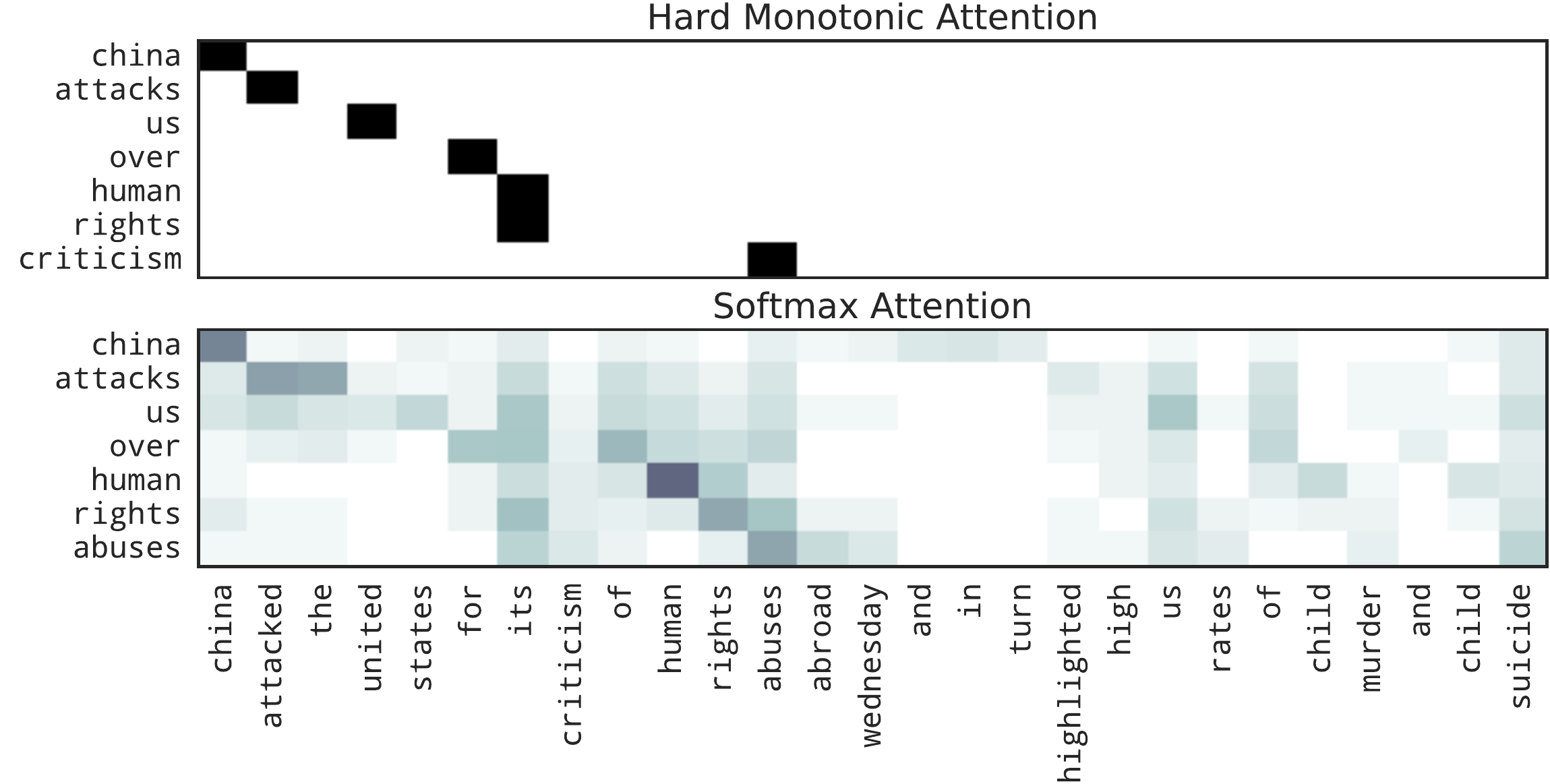}}
\caption{Additional example sentence-summary pair and attention alignment matrices for our hard monotonic model and the $\softmax$-based attention model of \citeappendix{liu2016text}.
The ground-truth summary is ``china attacks us human rights''.}
\label{fig:summarization_alignments_2}
\end{center}
\vskip -0.4in
\end{figure*}

\section{Monotonic Attention Distribution}
\label{sec:mono_derivation}

\begin{figure*}[t]
\vskip 0.2in
\begin{center}
\centerline{\includegraphics[width=\textwidth]{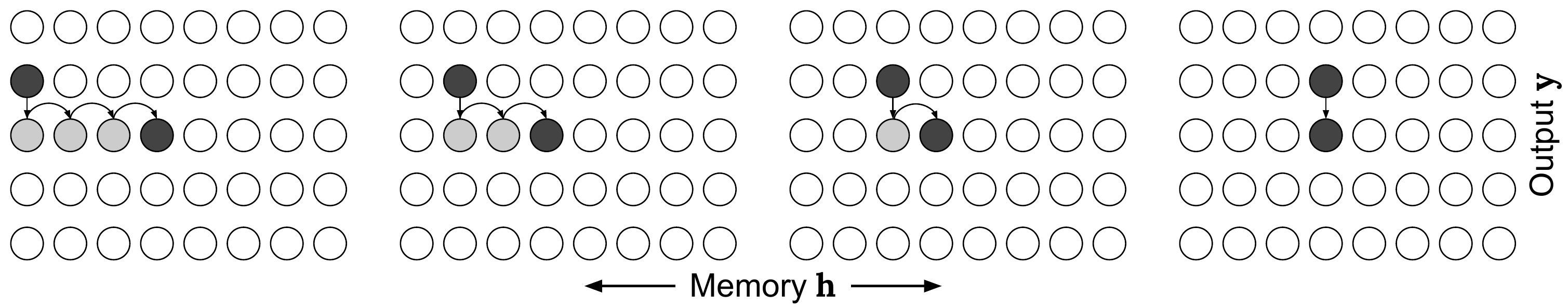}}
\caption{Visualization of \cref{eq:alpha_ij_2}.
In this example, we are showing the computation of $\alpha_{3, 4}$.
Each grid shows each of the four terms in the summation, corresponding to the possibilities that we attended to memory item $k = 1, 2, 3, 4$ at the previous output timestep $i - 1 = 2$.
Gray nodes with curved arrows represent the probability of not selecting to the $l$th memory entry ($1 - p_{i, l}$).
The black nodes represent the possibility of attending to memory item $j$ at timestep $i$.}
\label{fig:possible_paths}
\end{center}
\vskip -0.4in
\end{figure*}

Recall that our goal is to compute the expected value of $c_i$ under the stochastic process defined by \cref{eq:hard_energy,eq:hard_sigmoid,eq:z_ij}.
To achieve this, we will derive an expression for the probability that $c_i = h_j$ for $j \in \{1, \ldots, T\}$, which in accordance with \cref{eq:softmax} we denote $\alpha_{i, j}$.
For $i = 1$, $\alpha_{1, j}$ is the probability that memory element $j$ was chosen ($p_{1, j}$) multiplied by the probability that memory elements $k \in \{1, 2, \ldots, j - 1\}$ were not chosen ($(1 - p_{i, k})$), giving
\begin{equation}
        \alpha_{1, j} = p_{1, j}\prod_{k = 1}^{j - 1}(1 - p_{1, k}) \label{eq:alpha_1j}
\end{equation}
For $i > 0$, in order for $c_i = h_j$ we must have that $c_{i - 1} = h_k$ for some $k \in \{1, \ldots, j\}$ (which occurs with probability $\alpha_{i - 1, k}$) and that none of $h_k, \ldots, h_{j - 1}$ were chosen.
Summing over possible values of $k$, we have
\begin{equation}
        \alpha_{i, j} = p_{i, j} \sum_{k = 1}^j \left( \alpha_{i - 1, k} \prod_{l = k}^{j - 1} (1 - p_{i, l}) \right) \label{eq:alpha_ij_2}
\end{equation}
where for convenience we define $\prod_{n}^m x = 1$ when $n > m$.
We provide a schematic and explanation of \cref{eq:alpha_ij_2} in \cref{fig:possible_paths}.
Note that we can recover \cref{eq:alpha_1j} from \cref{eq:alpha_ij_2} by defining the special case $\alpha_{0, j} = \delta_j$ (i.e.\ $\alpha_{0, 1} = 1$ and $\alpha_{0, j} = 0$ for $j \in \{2, \ldots, T\}$).
Expanding \cref{eq:alpha_ij_2} reveals we can compute $\alpha_{i, j}$ directly given $\alpha_{i - 1, j}$ and $\alpha_{i, j - 1}$:
\begin{align}
        \alpha_{i, j} &= \sq{p_{i, j}\Bigg(\sum_{k = 1}^{j - 1} \left( \alpha_{i - 1, k} \prod_{l = k}^{j - 1} (1 - p_{i, l}) \right) + \alpha_{i - 1, j}\Bigg)} \\
                      &= p_{i, j}\Bigg((1 - p_{i, j - 1})\sum_{k = 1}^{j - 1} \left( \alpha_{i - 1, k} \prod_{l = k}^{j - 2} (1 - p_{i, l}) \right) + \alpha_{i - 1, j} \Bigg) \\
                      &= p_{i, j}\left((1 - p_{i, j - 1})\frac{\alpha_{i, j - 1}}{p_{i, j - 1}} + \alpha_{i - 1, j}\right) \label{eq:recurrence_relation_2}
\end{align}

Defining $q_{i, j} = \alpha_{i, j}/p_{i, j}$ produces \cref{eq:mono_q,eq:mono_masked_probs}.
\Cref{eq:recurrence_relation_2} also has an intuitive interpretation:
The expression $(1 - p_{i, j - 1})\alpha_{i, j - 1}/p_{i, j - 1}$ represents the probability that the model attended to memory item $j - 1$ at output timestep $i$, adjusted for the fact that memory item $j - 1$ was not chosen by multiplying $(1 - p_{i, j - 1})$ and dividing $p_{i, j - 1}$.
Adding $\alpha_{i - 1, j}$ reflects the additional possibility that the model attended to memory item $j$ at the previous output timestep, and multiplying by $p_{i, j}$ enforces that memory item $j$ was chosen at the current output timestep $i$.

\subsection{Recurrence Relation Solution}
\label{sec:recurrence}

While \cref{eq:recurrence_relation,eq:recurrence_relation_2} allow us to compute $\alpha_{i, j}$ directly from $\alpha_{i - 1, j}$ and $\alpha_{i, j - 1}$, the dependence on $\alpha_{i, j - 1}$ means that we must compute the terms $\alpha_{i, 1}, \alpha_{i, 2}, \ldots, \alpha_{i, T}$ sequentially.
This is in contrast to $\softmax$ attention, where these terms can be computed in parallel because they are independent.
Fortunately, there is a solution to the recurrence relation of \cref{eq:recurrence_relation} which allows the terms of $\alpha_i$ to be computed directly via parallelizable cumulative sum and cumulative product operations.
Using \cref{eq:mono_q} which substitutes $q_{i, j} = \alpha_{i, j}/p_{i, j}$, we have

\begin{align}
        q_{i, j} &= (1 - p_{i, j - 1})q_{i, j - 1} + \alpha_{i - 1, j} \\
        q_{i, j} - (1 - p_{i, j - 1})q_{i, j - 1} &= \alpha_{i - 1, j} \\
        \frac{q_{i, j}}{\prod_{k = 1}^{j} (1 - p_{i, k - 1})} - \frac{(1 - p_{i, j - 1})q_{i, j - 1}}{\prod_{k = 1}^{j} (1 - p_{i, k - 1})} &= \frac{\alpha_{i - 1, j}}{\prod_{k = 1}^{j} (1 - p_{i, k - 1})}  \\
        \frac{q_{i, j}}{\prod_{k = 1}^{j} (1 - p_{i, k - 1})} - \frac{q_{i, j - 1}}{\prod_{k = 1}^{j - 1} (1 - p_{i, k - 1})} &= \frac{\alpha_{i - 1, j}}{\prod_{k = 1}^{j} (1 - p_{i, k - 1})} \\
        \sum_{l = 1}^{j} \left( \frac{q_{i, l}}{\prod_{k = 1}^{l} (1 - p_{i, k - 1})} - \frac{q_{i, l - 1}}{\prod_{k = 1}^{l - 1} (1 - p_{i, k - 1})}\right) &= \sum_{l = 1}^{j}\frac{\alpha_{i - 1, l}}{\prod_{k = 1}^{l} (1 - p_{i, k - 1})} \\
        \frac{q_{i, j}}{\prod_{k = 1}^{j} (1 - p_{i, k - 1})} - q_{i, 0} &= \sum_{l = 1}^{j}\frac{\alpha_{i - 1, l}}{\prod_{k = 1}^{l} (1 - p_{i, k - 1})} \\
        q_{i, j} &= \left( \prod_{k = 1}^j (1 - p_{i, k - 1}) \right) \left(\sum_{l = 1}^{j}\frac{\alpha_{i - 1, l}}{\prod_{k = 1}^{l} (1 - p_{i, k - 1})}\right) \label{eq:cum_math} \\
        \Rightarrow q_i &= \texttt{cumprod}(1 - p_i)\texttt{cumsum}\left(\frac{\alpha_{i - 1}}{\texttt{cumprod}(1 - p_i)}\right)
\end{align}

where $\texttt{cumprod}(x) = [1, x_1, x_1x_2, \ldots, \prod_i^{|x| - 1} x_i]$ and $\texttt{cumsum}(x) = [x_1, x_1 + x_2, \ldots, \sum_i^{|x|} x_i]$.
Note that we use the ``exclusive'' variant of \texttt{cumprod}\footnote{This can be computed e.g. in Tensorflow via \texttt{tf.cumprod(x, exclusive=True)}} in keeping with our defined special case $p_{i, 0} = 0$.
Unlike the recurrence relation of \cref{eq:recurrence_relation}, these operations can be computed efficiently in parallel \citeappendix{ladner1980parallel}.
The primary disadvantage of this approach is that the product in the denominator of \cref{eq:cum_math} can cause numerical instabilities; we address this in \cref{sec:practitioners}.

\twocolumn

\section{Experiment Details}
\label{sec:experiment_details}

In this section, we give further details into the models and training procedures used in \cref{sec:experiments}.
Any further questions about implementation details should be directed to the corresponding author.
All models were implemented with TensorFlow \citeappendix{abadi2016tensorflow}.

\subsection{Speech Recognition}

\subsubsection{TIMIT}

Mel filterbank features were standardized to have zero mean and unit variance across feature dimensions according to their training set statistics and were fed directly into an RNN encoder with three unidirectional LSTM layers, each with 512 hidden units.
After the first and second LSTM layers, we downsampled hidden state sequences by skipping every other state before feeding into the subsequent layer.
For the decoder, we used a single unidirectional LSTM layer with 256 units, fed directly into the output $\softmax$ layer.
All weight matrices were initialized uniformly from $[-0.075, 0.075]$.
The output tokens were embedded via a learned embedding matrix with dimensionality 30, initialized uniformly from  $[-\sqrt{3/30}, \sqrt{3/30}]$.
Our decoder attention energy function used a hidden dimensionality of 512, with the scalar bias $r$ initialized to -1.
The model was regularized by adding weight noise with a standard deviation of $0.5$ after 2,000 training updates.
L2 weight regularization was also applied with a weight of $10^{-6}$.

We trained the network using Adam \citeappendix{kingma2014adam}, with $\beta_1 = 0.9$, $\beta_2 = 0.999$, and $\epsilon = 10^{-6}$.
Utterances were fed to the network with a minibatch size of 4.
Our initial learning rate was $10^{-4}$, which we halved after 40,000 training steps.
We clipped gradients when their global norm exceeded 2.
We used three training replicas.
Beam search decoding was used to produce output sequences with a beam width of 10.

\subsubsection{Wall Street Journal}
\label{sec:wsj_details}

The input 80 mel filterbank / delta / delta-delta features were organized as a $T \times 80 \times 3$ tensor, i.e.\ raw features, deltas, and delta-deltas are concatenated along the ``depth'' dimension.
This was passed into a stack of two convolutional layers with ReLU activations, each consisting of 32 $3 \times 3 \times $ depth kernels in time $\times$ frequency.
These were both strided by $2 \times 2$ in order to downsample the sequence in time, minimizing the computation performed in the following layers.
Batch normalization \citeappendix{ioffe2015batch} was applied prior to the ReLU activation in each layer.
All encoder weight matrices and filters were initialized via a truncated Gaussian with zero mean and a standard deviation of 0.1.

This downsampled feature sequence was then passed into a single unidirectional convolutional LSTM layer using 1x3 filter (i.e.\ only convolving across the frequency dimension within each timestep).
Finally, this was passed into a stack of three unidirectional LSTM layers of size 256, interleaved with a 256 dimensional linear projection, following by batch normalization, and a ReLU activation.
Decoder weight matrices were initialized uniformly at random from $[-0.1, 0.1]$.

The decoder input is created by concatenating a 64 dimensional embedding corresponding to the symbol emitted at the previous timestep, and the 256 dimensional attention context vector.
The embedding was initialized uniformly from $[-1, 1]$.
This was passed into a single unidirectional LSTM layer with 256 units.
We used an attention energy function hidden dimensionality of 128 and initialized the bias scalar $r$ to -4.
Finally the concatenation of the attention context and LSTM output is passed into the $\softmax$ output layer.

We applied label smoothing \citeappendix{chorowski2017towards}, replacing $\hat{y}_t$, the target at time $t$, with $(0.015\hat{y}_{t - 2} + 0.035\hat{y}_{t - 1} + \hat{y}_t + 0.035\hat{y}_{t + 1} + 0.015\hat{y}_{t + 2})/1.1$.
We used beam search decoding at test time with rank pruning at 8 hypotheses and a pruning threshold of 3.

The network was trained using teacher forcing on minibatches of 8 input utterances, optimized using Adam \citeappendix{kingma2014adam} with $\beta_1= 0.9$, $\beta_2 = 0.999$, and $\epsilon = 10^{-6}$.
Gradients were clipped to a maximum global norm of 1.
We set the initial learning rate to $0.0002$ and decayed by a factor of 10 after 700,000, 1,000,000, and 1,300,000 training steps.
L2 weight decay is used with a weight of $10^{-6}$, and, beginning from step 20,000, Gaussian weight noise with standard deviation of 0.075 was added to weights for all LSTM layers and decoder embeddings.
We trained using 16 replicas.

\subsection{Sentence Summarization}

For data preparation, we used the same Gigaword data processing scripts provided in \citeappendix{rush2015neural} and tokenized into words by splitting on spaces.
The vocabulary was determined by selecting the most frequent 200,000 tokens.
Only the tokens of the first sentence of the article were used as input to the model.
An embedding layer was used to embed tokens into a 200 dimensional space; embeddings were initialized using random normal distribution with mean $0$ and standard deviation $10^{-4}$.

We used a 4-layer bidirectional LSTM encoder with 4 layers and a single-layer unidirectional LSTM decoder.
All LSTMs, and the attention energy function, had a hidden dimensionality of 256.
The decoder LSTM was fed directly into the $\softmax$ output layer.
All weights were initialized uniform-randomly between $-0.1$ and $0.1$.
In our monotonic alignment decoder, we initialized $r$ to -4.
At test time, we used a beam search over possible label sequences with a beam width of 4.

A batch size of 64 was used and the model was trained to minimize the sampled-$\softmax$ cross-entropy loss with 4096 negative samples.
The Adam optimizer \citeappendix{kingma2014adam} was used with $\beta_1= 0.9$, $\beta_2 = 0.999$, and $\epsilon = 10^{-4}$,  and an initial learning rate of $10^{-3}$; an exponential decay was applied by multiplying the initial learning rate by $.98^{n/30000}$ where $n$ is the current training step.
Gradients were clipped to have a maximum global norm of 2.
Early stopping was used with respect to validation loss and took about 300,000 steps for the baseline model, and 180,000 steps for the monotonic model.
Training was conducted on 16 machines with 4 GPUs each.
We reported ROUGE scores computed over the test set of \citeappendix{rush2015neural}.

\subsection{Machine Translation}

Overall, we followed the model of \citeappendix{luong2015stanford} closely; our hyperparameters are largely the same:
Words were mapped to 512-dimensional embeddings, which were learned during training.
We passed sentences to the network in minibatches of size 128.
As mentioned in the text, we used two unidirectional LSTM layers in both the encoder and decoder.
All LSTM layers, and the attention energy function, had a hidden dimensionality of 512.
We trained with a single replica for 40 epochs using Adam \citeappendix{kingma2014adam} with $\beta_1= 0.9$, $\beta_2 = 0.999$, and $\epsilon = 10^{-8}$.
We performed grid searches over initial learning rate and decay schedules separately for models using each of the two energy functions \cref{eq:modified_energy} and \cref{eq:luong_energy}.
For the model using \cref{eq:modified_energy}, we used an initial learning rate of $0.0005$, and after $10$ epochs we multiplied the learning rate by $0.8$ each epoch; for \cref{eq:luong_energy} we started at $0.001$ and multiplied by $0.8$ each epoch starting at the eighth epoch.
Parameters were uniformly initialized in range $[-0.1, 0.1]$.
Gradients were scaled whenever their norms exceeded 5.
We used dropout with probability $0.3$ as described in \citeappendix{pham2014dropout}.
Unlike \citeappendix{luong2015stanford}, we did not reverse source sentences in our monotonic attention experiments.
We set $r = -2$ for the attention energy function bias scalar for both \cref{eq:modified_energy} and \cref{eq:luong_energy}.
We used greedy decoding (i.e.\ no beam search) at test time.

\section{Future Work}
\label{sec:future_work}

We believe there are a variety of promising extensions of our monotonic attention mechanism, which we outline briefly below.
\begin{itemize}[leftmargin=*,topsep=0pt,itemsep=-1ex,partopsep=2ex,parsep=2ex]
\item The primary drawback of training in expectation is that it retains the quadratic complexity during training.
One idea would be to replace the cumulative product in \cref{eq:alpha_ij} with the thresholded remainder method of \citeappendix{graves2016adaptive} and \citeappendix{grefenstette2015learning}, but in preliminary experiments we were unable to successfully learn alignments with this approach.
Alternatively, we could further our investigation into gradient estimators for discrete decisions (such as REINFORCE or straight-through) instead of training in expectation \citeappendix{bengio2013estimating}.
\item As we point out in \cref{sec:energy}, our method can fail when the attention energies $e_{i, j}$ are poorly scaled.
This primarily stems from the strict enforcement of monotonicity.
One possibility to mitigate this would be to instead regularize the model with a soft penalty which discourages non-monotonic alignments, instead of preventing them outright.
\item In some problems, the input-output alignment is non-monotonic only in small regions.
A simple modification to our approach which would allow this would be to subtract a constant integer from $t_{i - 1}$ between output timesteps.
Alternatively, utilizing multiple monotonic attention mechanisms in parallel would allow the model to attend to disparate memory locations at each output timestep (effectively allowing for non-monotonic alignments) while still maintaining linear-time decoding.
\item To facilitate comparison, we sought to modify the standard $\softmax$-based attention framework as little as possible.
As a result, we have thus far not fully taken advantage of the fact that the decoding process is much more efficient.
Specifically, the attention energy function of \cref{eq:standard_energy} was primarily motivated by the fact that it is trivially parallelizable so that its repeated application is inexpensive.
We could instead use a recurrent attention energy function, whose output depends on both the attention energies for prior memory items and those at the previous output timestep.
\end{itemize}

\section{How much faster is linear-time decoding?}
\label{sec:faster}

Throughout this paper, we have emphasized that one advantage of our approach is that it allows for linear-time decoding, i.e.\ the decoder RNN only makes a single pass over the memory in the course of producing the output sequence.
However, we have thus far not attempted to quantify how much of a speedup this incurs in practice.
Towards this end, we conducted an additional experiment to measure the speed of efficiently-implemented $\softmax$-based and hard monotonic attention mechanisms.
We chose to focus solely on the speed of the attention mechanisms rather than an entire RNN sequence-to-sequence model because models using these attention mechanisms are otherwise equivalent.
Measuring the speed of the attention mechanisms alone allows us to isolate the difference in computational cost between the two approaches.

Specifically, we implemented both attention mechanisms using the highly-efficient C++ linear algebra package Eigen \citeappendix{guennebaud2010eigen}.
We set entries of the memory $\mathbf{h}$ and the decoder hidden states $s_i$ to random vectors with entries sampled uniformly in the range $[-1, 1]$.
We then computed context vectors following \cref{eq:softmax,eq:softmax_context} for the $\softmax$ attention mechanism and following \cref{alg:hard_monotonic} for hard monotonic attention.
We varied the input and output sequence lengths and averaged the time to produce all of the corresponding context vectors over 100 trials for each setting.

\begin{figure}[t]
\vskip 0.2in
\begin{center}
\centerline{\includegraphics[width=\columnwidth]{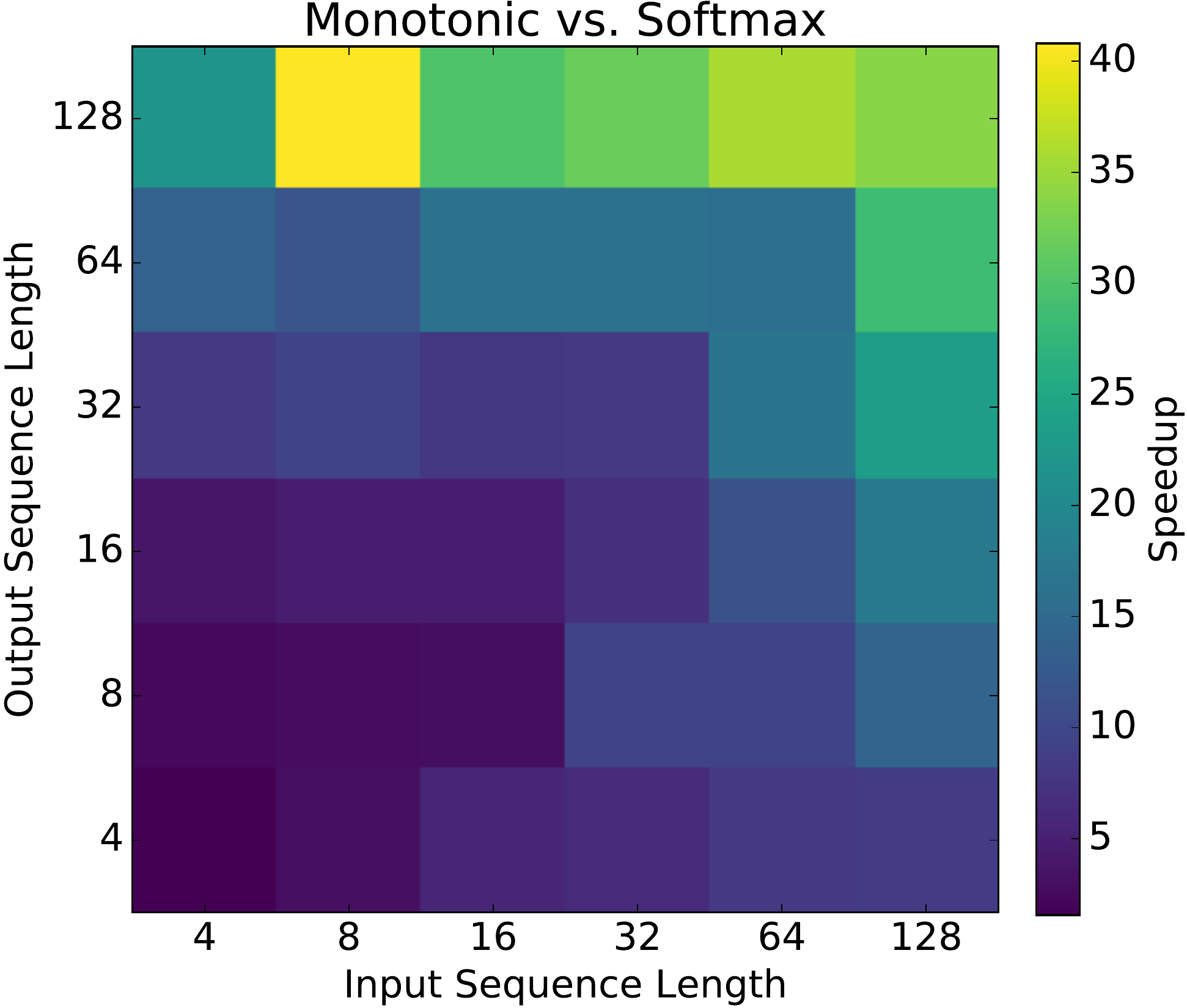}}
\caption{Speedup of hard monotonic attention mechanism compared to $\softmax$ attention on a synthetic benchmark.}
\label{fig:speed}
\end{center}
\vskip -0.2in
\end{figure}

The speedup of the monotonic attention mechanism compared to $\softmax$ attention is visualized in \cref{fig:speed}.
We found monotonic attention to be about $4-40\times$ faster depending on the input and output sequence lengths.
The most prominent difference occurred for short input sequences and long output sequences; in these cases the monotonic attention mechanism finishes processing the input sequence before it finishes producing the output sequence and therefore is able to stop computing context vectors.
We emphasize that these numbers represent the best-case speedup from our approach; a more general insight is simply that our proposed hard monotonic attention mechanism has the potential to make decoding significantly more efficient for long sequences.
Additionally, this advantage is distinct from the fact that our hard monotonic attention mechanism can be used for online sequence-to-sequence problems.
We also emphasize that at training time, we expect our soft monotonic attention approach to have roughly the computational cost as standard $\softmax$ attention, thanks to the fact that we can compute the resulting attention distribution in parallel as described in \cref{sec:recurrence}.
The code used for this benchmark is available in the repository for this paper.\footnote{\url{https://github.com/craffel/mad}}

\section{Practitioner's Guide}
\label{sec:practitioners}

Because we are proposing a novel attention mechanism, we share here some insights gained from applying it in various settings in order to help practitioners try it on their own problems:
\begin{itemize}[leftmargin=*,topsep=0pt,itemsep=-1ex,partopsep=2ex,parsep=2ex]
\item The recursive structure of computing $\alpha_{i, j}$ in \cref{eq:alpha_ij} can result in exploding gradients.
We found it vital to apply gradient clipping in all of our experiments, as described in \cref{sec:experiment_details}.
\item Many automatic differentiation packages can produce numerically unstable gradients when using their cumulative product function.\footnote{\url{https://github.com/tensorflow/tensorflow/issues/3862}}\footnote{\url{https://github.com/Theano/Theano/issues/5197}}
Our simple solution was to compute the product in log-space, i.e.\ replacing $\prod_n x_n = \exp(\sum_i\log(x_n))$.
\item In addition, the product in the denominator of \cref{eq:cum_math} can become negligibly small because the terms $(1 - p_{i, k - 1})$ all fall in the range $[0, 1]$.
The simplest way to prevent the resulting numerical instabilities is to clip the range of the denominator to be within $[\epsilon, 1]$ where $\epsilon$ is a small constant (we used $\epsilon = 10^{-10}$).
This can result in incorrect values for $\alpha_{i, j}$ particularly when some $p_{i, j}$ are close to 1, but we encountered no discernible effect on our results.
\item Alternatively, we found in preliminary experiments that simply setting the denominator to $1$ still produced good results.
This can be explained by the observation that when all $p_{i, j} \in \{0, 1\}$ (which we encourage during training), \cref{eq:cum_math} is equivalent to the recurrence relation of \cref{eq:recurrence_relation} even when the denominator is $1$.
\item As we mention in the experiment details of the previous section, we ended up using a small range of values for the initial energy function scalar bias $r$.
In general, performance was not very sensitive to this parameter, but we found small performance gains from using values in $\{-5, -4, -3, -2, -1\}$ for different problems.
\item More broadly, while the attention energy function modifications described in \cref{sec:energy} allowed models using our mechanism to be effectively trained on all tasks we tried, they were not always necessary for convergence.
Specifically, in speech recognition experiments the performance of our model was the same using \cref{eq:standard_energy} and \cref{eq:modified_energy}, but for summarization experiments the models were unable to learn to utilize attention when using \cref{eq:standard_energy}.
For ease of implementation, we recommend starting with the standard attention energy function of \cref{eq:standard_energy} and then applying the modifications of \cref{eq:modified_energy} if the model fails to utilize attention.
\item It is occasionally recommended to reverse the input sequence prior to feeding it into sequence-to-sequence models \citeappendix{sutskever2014sequence}.
This violates our assumption that the input should be processed in a left-to-right manner when computing attention, so should be avoided.
\item Finally, we highly recommend visualizing the attention alignments $\alpha_{i, j}$ over the course of training.
Attention provides valuable insight into the model's behavior, and failure modes can be quickly spotted (e.g. if $\alpha_{i, j} = 0$ for all $i$ and $j$).
\end{itemize}
With the above factors in mind, on all problems we studied, we were able to replace $\softmax$-based attention with our novel attention and immediately achieve competitive performance.

\bibliographystyleappendix{icml2017}
\bibliographyappendix{example_paper}

\end{document}